\documentclass{article}
\usepackage{spconf,amsmath,graphicx,hyperref}

\usepackage{xspace,xcolor}

\makeatletter
\DeclareRobustCommand\onedot{\futurelet\@let@token\@onedot}
\def\@onedot{\ifx\@let@token.\else.\null\fi\xspace}

\def\eg{\emph{e.g}\onedot}

\makeatother

\newcommand{\ours}{\textsc{MedFact-R1}}

\usepackage[numbers]{natbib}
\usepackage{booktabs}
\usepackage{multirow}
\usepackage{xcolor,colortbl}
\usepackage[capitalize,noabbrev]{cleveref}
\usepackage{soul}
\usepackage{amssymb}
\usepackage{pifont}

\crefname{table}{Tab.}{Tabs.}
\crefname{figure}{Fig.}{Figs.}
\crefname{section}{Sec.}{Secs.}

\definecolor{tabfirst}{rgb}{1, 0.7, 0.7} 
\definecolor{tabsecond}{rgb}{1, 0.85, 0.7} 
\definecolor{myblue}{HTML}{4A90E2}

\newcommand{\logopic}{%
   \raisebox{-0.5ex}{
      \includegraphics[height=2.5ex]{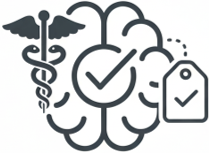}
  }%
}

\makeatletter
\renewcommand{\@makefnmark}{\hbox{\@textsuperscript{\fnsymbol{footnote}}}}
\makeatother

\title{MedFact-R1: Towards Factual Medical Reasoning via Pseudo-Label Augmentation}

\name{Gengliang LI$^{1,6\dagger}$, Rongyu CHEN$^{2,5\dagger}$, Bin LI$^3$, Linlin YANG$^{4\textasteriskcentered}$, Guodong DING$^2$}

\address{$^1$Baosight, $^2$NUS, $^3$SIAT, CAS, $^4$CUC, $^5$Microsoft, $^6$ANU}
\begin{document}
%
\maketitle

\begin{abstract}
Ensuring factual consistency and reliable reasoning remains a critical challenge for medical vision-language models.  
We introduce \ours, a two-stage framework that integrates external knowledge grounding with reinforcement learning to improve the factual medical reasoning.
The first stage uses pseudo-label supervised fine-tuning (SFT) to incorporate external factual expertise;
while the second stage applies Group Relative Policy Optimization (GRPO) with four tailored factual reward signals to encourage self-consistent reasoning.
Across three public medical QA benchmarks, \ours\ delivers up to $\mathbf{22.5\%}$ absolute improvement in factual accuracy over previous state-of-the-art methods.
Ablation studies highlight the necessity of pseudo-label SFT cold start and validate the contribution of each GRPO reward, underscoring the synergy between knowledge grounding and RL-driven reasoning for trustworthy medical AI.
Codes are released at \href{https://github.com/Garfieldgengliang/MEDFACT-R1}{https://github.com/Garfieldgengliang/MEDFACT-R1}.
\end{abstract}
\vspace{-0.2cm}
\section{Introduction}\label{sec:intro}
\footnotetext{$^{\dagger}$These authors contributed equally to this work.}
\footnotetext{$^{*}$Corresponding author.}
Medical diagnosis represents one of the most critical frontiers in signal processing and machine learning, embodying the vision of technology serving humanity.
It demands exceptional expertise, stringent accuracy, and the ability to reason over inherently complex data.
Yet, progress is hindered by the scarcity of high-quality diagnostic data, restricted by both professional requirements and privacy constraints.
These limitations complicate model training and often result in unreliable behavior -  such as reliance on spurious correlations, misjudgments, and missed diagnoses.  
In the medical domain, where decisions directly affect human lives, such errors are unacceptable.   
Overcoming these challenges is essential for developing deep learning systems that can achieve reliable performance in real-world clinical practice~\cite{rule24emnlp_unc}.   

Recently, large-scale Vision-Language Models (VLMs)~\cite{gpt4_23_openai,qwen25-vl25_alibaba,ds-r125_ds} have advanced rapidly, transforming various industries.  Their extension to the medical domain has shown promising potential, with recent efforts~\cite{llava-med23neurips_msft} curating medical datasets and fine-tuned VLMs for specialized applications.  
However, factual reliability remains a major obstacle: existing models often generate hallucinations and factual errors in high-stake scenarios. 
To address this, RULE~\cite{rule24emnlp_unc} introduces risk-controlled Retrieval-Augmented Generation (RAG), which balances external retrieval and internal knowledge, yielding notable factuality improvements.  

Concurrently, many post-training efforts have successfully exploited the knowledge and potential of vision-language models themselves, among which advanced Reinforcement Learning (RL) has emerged as a prominent example.
Unlike supervised learning via next-token prediction, reinforcement learning optimizes task policies using reward signals without relying on detailed annotations.
GRPO~\cite{grpo24_ds,sophiavl-r1_25_cuhk}, one of the most advanced RL post-training methods, surpasses supervised fine-tuning in generalization by unlocking ``aha'' moments in reasoning~\cite{ds-r125_ds}, which sets it apart from prior approaches such as PPO~\cite{ppo17_openai}, DPO~\cite{dpo23neurips_stanford} and traditional ones~\cite{miura21naaclw_stanford,delbrouck22emnlp_stanford}.
Despite the impressive results, sufficient domain knowledge is found essential to prevent unrealistic or verbose outputs, which often rely on the costly curation of diverse annotated medical datasets~\cite{mmedagent-rl25_unc,medground-r1_25miccai_shailab}.

\begin{figure*}[!th]
  \centering
  \includegraphics[width=0.95\linewidth]{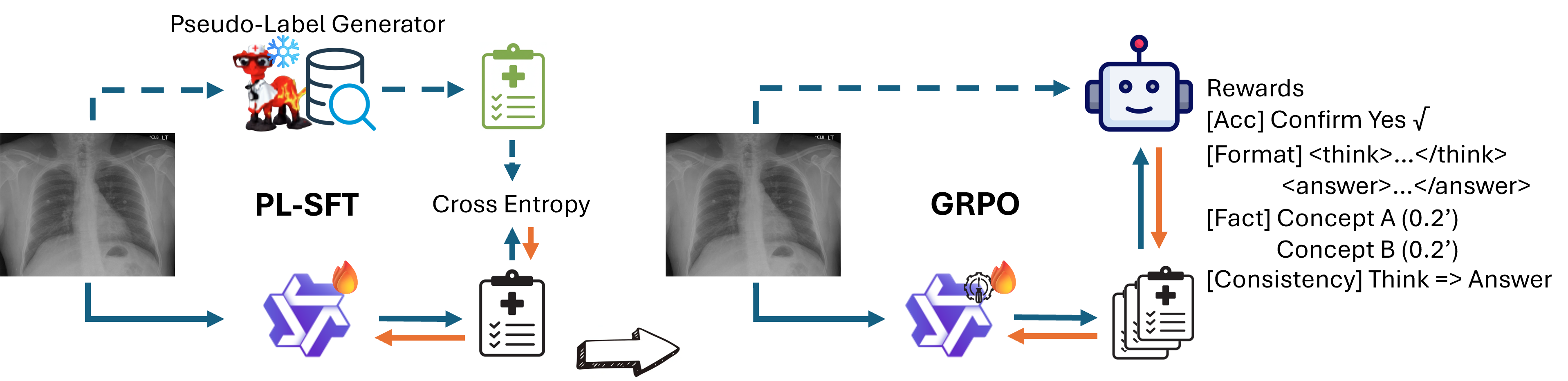}

  \caption{\textbf{Overview of our method \ours\ consisting of two stages, Pseudo-Label Supervised Fine-Tuning \& GRPO-based Reinforcement Learning.}
  The \textbf{\color{myblue}blue} and \textbf{\color{orange}orange} arrows represent the feedforward and backpropagation flow, respectively}

  \label{fig:ov}
\end{figure*}

To address the challenges of data scarcity in RL and reliance on external knowledge, we propose \logopic\ours\ (\cref{fig:ov}).
It is a two-stage framework that combines SFT on factual pseudo-diagnosis data with RL-incentivized reasoning to activate factual capabilities without external reference.
In the \emph{first} stage, we apply a generator with factuality risk control to generate pseudo-diagnosis data for SFT, thereby more effectively expanding medical knowledge exposure and strengthening the model’s foundation.  
In the \emph{second} stage, we adopt GRPO post-training to help the model fully digest and reflect on valuable diagnostic data.
To better align with the medical diagnosis scenario, we carefully design four reward components, considering answer correctness, formal presentation, contextual relevance, and self-consistency with factual reasoning.
Our GRPO encourages reasoning grounded in facts and allows the model to generalize beyond the observed cases and improve medical factuality. 

The experimental results reveal strong and consistent improvements across several public medical Question Answering benchmarks.  
\logopic\ours\ enhances all evaluation metrics, yielding more factual and reliable diagnostic outputs and validating the effectiveness of our framework in advancing the state-of-the-art medical factuality reasoning.  
In addition, our ablation studies provide deeper insights into the necessity of a pseudo-label SFT start and clarify the individual contributions of each factual reward.

In summary, our contributions are as follows:
\textbf{1)} We pioneer better integration of advanced reinforcement learning GRPO into medical QA via a two-stage training pipeline combining SFT and RL post-training paradigms to enhance the VLM's own factual reasoning;
\textbf{2)} We initialize training with {SFT} using factual \emph{Pseudo-labels} to effectively absorb external medical knowledge;
\textbf{3)} To address sparse reward signals and further improve factuality, \emph{factual rewards} are tailored for GRPO post-training, reducing hallucinations and promoting self-justification;
\textbf{4)} \logopic\ours\ sets a new state-of-the-art, achieving scores exceeding $95\%$ on all evaluation metrics across diverse medical QA benchmarks, with gains of up to $22.5\%$ over prior methods.

\vspace{-0.2cm}
\section{\ours}
\vspace{-0.2cm}
\subsection{Task Formulation}
\vspace{-0.1cm}
{Medical visual Question Answering (QA) is the task of answering medical questions based on images~\cite{llava-med23neurips_msft}. 
Formally, given an image $I$ and question $Q$, the model generates an answer in free-form text.
For VLMs, the first token of the generated answer is expected to be either ``\textit{yes}'' or ``\textit{no}'';
this token is mapped to a binary label $\hat{A} \in \{0, 1\}$, indicating clinically correct or incorrect response.}

\subsection{Supervised Fine-Tuning with Pseudo-Labels}\label{sec:sft}

We begin with Supervised Fine-Tuning (SFT) using next-token prediction under maximum likelihood estimation to establish a foundation for factual reasoning.
To this end, we generate pseudo-labels via calibrated retrieval to mitigate factuality risks and preference alignment to harmonize external priors with internal knowledge~\cite{rule24emnlp_unc}.
This strategy distills external medical knowledge into compact supervision signals that suppress hallucinations and reinforce the factual consistency.
\vspace{-0.3cm}
\subsection{GRPO Post-Training}\label{sec:rl}

Post-SFT, GRPO~\cite{grpo24_ds} guides policy updates via rule-based rewards on generated outputs using group-based Monte Carlo advantage estimation and policy gradient.
We design four complementary types of reward values for GRPO-based reinforcement learning, with the total reward given by the normalized sum of these components from generated answers:

\noindent\textbf{Accuracy Reward.}\label{sec:acc} 
We assess the correctness of the predicted answers by comparing them with pseudo-labels. For binary classification, the reward is 1.0 for exact matches and 0 otherwise.

\noindent\textbf{Format Reward.}\label{sec:format}
To encourage structured reasoning, we require the outputs to include a thought process enclosed within \texttt{<think>} and \texttt{</think>} tags and a concise final answer within \texttt{<answer>} and \texttt{</answer>} tags.
A reward of 1 is assigned only if all four tags appear exactly once and no extraneous content exists outside these regions; otherwise, the reward is 0.

\noindent\textbf{{Fact Reward.}}\label{sec:fact}
To promote factual alignment in medical reasoning, we design a reward based on the presence of clinically grounded concepts in the model output.
For each training question, a small set of domain-specific concepts is extracted using GPT-4~\cite{gpt4_23_openai}, serving as factual anchors.
For example, the question \textit{``Does the chest radiograph show any signs of lung infection or congestion?''} yields concepts such as \textit{``lung infection, congestion, chest radiography''}.
Each concept correctly reflected in the answer contributes 0.2 points, counted once.
This guides the model to express medically relevant knowledge, offering a targeted signal for factual grounding.

\begin{table*}[!th]
\centering
\begin{sc}
\resizebox{.95\linewidth}{!}{
\begin{tabular}{lccccccccccccc}
    \toprule
    \multirow{2}{*}{\textbf{Models}} & \multirow{2}{*}{\textbf{Venues}} & \multicolumn{4}{c}{\textbf{IU-Xray}} & \multicolumn{4}{c}{\textbf{Harvard-FairVLMed}} & \multicolumn{4}{c}{\textbf{MIMIC-CXR}} \\
    & & \textbf{Acc.} & \textbf{Pre.} & \textbf{Rec.} & \textbf{F1} & \textbf{Acc.} & \textbf{Pre.} & \textbf{Rec.} & \textbf{F1} & \textbf{Acc.} & \textbf{Pre.} & \textbf{Rec.} & \textbf{F1} \\
    \midrule
    LLaVA-Med v1.5 (7B)~\cite{llava-med23neurips_msft} & arXiv'24 & 75.47 & 53.17 & 80.49 & 64.04 & 63.03 & 92.13 & 61.46 & 74.11 & 75.79 & 81.01 & 79.38 & 80.49 \\ 
    \quad + Greedy & - & 76.88 & 54.41 & 82.53 & 65.59 & 78.32 & 91.59 & 82.38 & 86.75 & {82.54} & 82.68 & 81.73 & 85.98 \\
    \quad + Beam Search & - & 76.91 & 54.37 & {84.13} & 66.06 & {80.93} & {93.01} & {82.78} & {88.08} & 81.56 & {83.04} & {84.76} & {86.36} \\
    \quad + DoLa & - & {78.00} & {55.96} & 82.69 & {66.75} & 76.87 & 92.69 & 79.40 & 85.53 & 81.35 & 80.94 & 81.07 & 85.73 \\
    \quad + OPERA & - & 70.59 & 44.44 & \cellcolor{tabfirst}{100.0} & 61.54 & 71.41 & 92.72 & 72.49 & 81.37 & 69.34 & 72.04 & 79.19 & 76.66 \\
    \quad + VCD & - & 68.99 & 44.77 & 69.14 & 54.35 & 65.88 & 90.93 & 67.07 & 77.20 & 70.89 & 78.06 & 73.23 & 75.57 \\
        \midrule
    \quad MedDr~\cite{meddr24_hkust} & arXiv'24 & 83.33 & - & - & 67.80 & 70.17 & - & - & 80.72 & 55.16 & - & - & 56.18 \\
    \quad RULE~\cite{rule24emnlp_unc} & EMNLP'24 & 87.84 & 75.41 & 80.79 & 78.00 & 87.12 & 93.57 & 96.69 & 92.89 & 83.92 & 87.01 & 82.89 & 87.49 \\
    \quad MMed-RAG~\cite{mmed-rag25iclr_unc} & ICLR'25 & {89.54} & - & - & {80.72} & 87.94 & - & - & 92.78 & 83.57 & - & - & {88.49} \\
    \quad FactMM-RAG~\cite{factmm-rag25naacl_cmu} & NAACL'25 & 84.51 & - & - & 68.51 & 83.67 & - & - & 87.21 & 77.58 & - & - & 81.86 \\
    \midrule
    Qwen2.5-VL-3B~\cite{qwen25-vl25_alibaba} & arXiv'25 & 61.21 & 37.32 & 66.91 & 47.91 & 42.83 & 85.83 & 37.32 & 52.00 & 53.57 & 80.45 & 30.81 & 44.53 \\ 
    \quad \textbf{Ours} & - & \cellcolor{tabfirst}97.63 & \cellcolor{tabfirst}95.62 & 95.48 & \cellcolor{tabfirst}95.55 & \cellcolor{tabfirst}96.54 & \cellcolor{tabfirst}96.17 & \cellcolor{tabfirst}99.97 & \cellcolor{tabfirst}98.03 & \cellcolor{tabfirst}95.36 & \cellcolor{tabfirst}93.13 & \cellcolor{tabfirst}99.91 & \cellcolor{tabfirst}96.40 \\
    \bottomrule
\end{tabular}
}
\end{sc}
\caption{\textbf{Comparisons with the state-of-the-art across three medical benchmarks.}
The best results are colored in \colorbox{tabfirst}{red}.}
\label{tab:sotas}

\vspace{-0.2cm}
\end{table*}

\noindent\textbf{Consistency Reward.}\label{sec:cons}
To reinforce factual and logical coherence in clinical contexts, we introduce a consistency reward that evaluates the alignment between the reasoning and final answer.
We use GPT-4 to assess contextual consistency, referencing curated examples of medically sound and unsound outputs.
A reward of 1 is given if the answer is logically supported by its reasoning, particularly in terms of clinical interpretation; otherwise, a penalty of -0.5 is applied.
This encourages the model to maintain internal consistency when drawing conclusions from the medical evidence.

\vspace{-0.3cm}
\section{Experiments}

\subsection{Datasets and Implementation}

\noindent\textbf{Datasets.} Experiments are conducted on the three medical benchmarks: \textbf{IU-Xray}~\cite{iu-xray16jamia}, \textbf{Harvard-FairVLMed}~\cite{harvard-fairvlmed24fairclip_harvard} and \textbf{MIMIC-CXR}~\cite{mimic-cxr19_mit}.
{{IU-Xray} consists of chest X-ray images paired with diagnostic reports, providing a set of 2,573 inference samples for evaluating image-text alignment and report generation.
{Harvard-FairVLMed} focuses on fairness assessment in multimodal fundus imaging, with 4,285 samples spanning diverse demographic and clinical scenarios.
{MIMIC-CXR} is a large-scale dataset of chest radiographs associated with free-text reports, with 3,470 samples curated for a comprehensive evaluation of factuality and reasoning.
We adopt the official splits and standardized evaluation protocols provided by~\cite{rule24emnlp_unc}.} 

\noindent\textbf{Metrics.} We evaluate using typical classification metrics, {Accuracy} (\textbf{Acc.}), Precision (\textbf{Pre.}), Recall (\textbf{Rec.}), and \textbf{F1} score.
{\emph{Accuracy} measures the proportion of correctly predicted samples, whereas \emph{Precision} and \emph{Recall} quantify the model’s ability to identify relevant instances and recover all true positive diseases, respectively.
The \emph{F1} score provides a harmonic mean of Precision and Recall, offering a balanced assessment of the model performance in scenarios with class imbalance.}

\noindent \textbf{Implementation.}
Our model is built upon \textsc{Qwen-2.5-VL-3B}, trained with the open-sourced GRPO framework$^{1}$\footnotetext{$^{1}$https://github.com/StarsfieldAI/R1-V}. We set the maximum prompt length to 8,192 and the maximum completion length to 2,048, allowing for long-context modeling.
The maximum image input size is set to 501,760 pixels.
We sample 6 generations per input to balance exploration and convergence with a learning rate of $5e^{-5}$ and keep the other hyperparameters as default.
The model is trained for 2 epochs with \texttt{bf16} mixed precision, gradient checkpointing, and {FlashAttention}~\cite{flashattn22neurips_stanford}.
Training takes 10 hours on 4 NVIDIA A100 GPUs 80 GB with a per-device batch size of 1 and gradient accumulation set to 1.

\subsection{Comparisons with the State-of-the-Art}

\begin{figure*}[!th]
  \centering
  \includegraphics[width=0.95\linewidth]{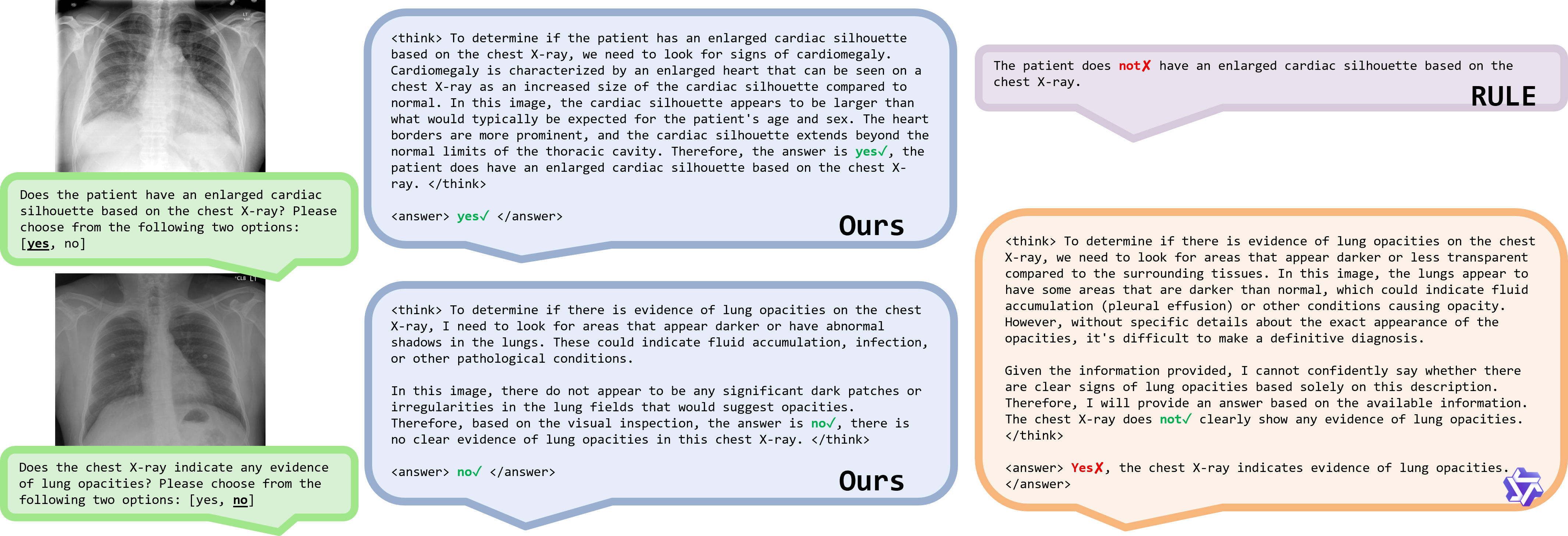}

  \caption{\textbf{The case analyses show enhanced factuality of our medical diagnosis compared to \textsc{RULE} and \textsc{Qwen2.5-VL}.}
  Correct answers are marked with {\color{green}\checkmark}, while incorrect ones are denoted by {\color{red}\ding{55}}.}

  \label{fig:exs}
\end{figure*}

As shown in \cref{tab:sotas}, the baselines include \textsc{Qwen2.5-VL-3B}~\cite{qwen25-vl25_alibaba} and \textsc{LLaVA-Med v1.5}~\cite{llava-med23neurips_msft} fine-tuned on medical data, along with various traditional post-hoc enhancements (\eg Greedy Search).
They exhibit variable performance across datasets, highlighting the challenges of consistent generalization.  
Among the SOTAs, the representative \textsc{RULE}~\cite{rule24emnlp_unc} addresses some of these limitations through calibrated retrieval and Direct Preference Optimization RL~\cite{dpo23neurips_stanford}, yet is constrained by reasoning depth~\cite{grpo24_ds}.  
While sharing the goal of enhancing factual reasoning with RL-based SOTAs, our method achieves this via factual pseudo-labels that facilitate GRPO, avoiding costly annotations and remaining compatible with them.
Notably, even with a smaller 3B model, our model significantly outperforms the larger 7B SOTA counterparts by over $10\%$ in accuracy across modalities and clinical domains, demonstrating the superiority of our overall framework.
The observed gains in both precision and recall suggest that our training strategy not only reduces hallucinations but also enhances the model’s ability to capture subtle clinical cues, thereby balancing sensitivity and specificity while advancing factual reliability and clinical interpretability.
Qualitative cases are shown in \cref{fig:exs}.

\subsection{Ablation Studies}

\subsubsection{Training Strategies}

\begin{table}[!th]
\centering
\begin{sc}
\resizebox{1\linewidth}{!}{
\begin{tabular}{ccccccccc}
    \toprule
    \multirow{2}{*}{PL-SFT} & \multicolumn{4}{c}{{GRPO}} & \multirow{2}{*}{\textbf{Acc.}} & \multirow{2}{*}{\textbf{Pre.}} & \multirow{2}{*}{\textbf{Rec.}} & \multirow{2}{*}{\textbf{F1}} \\
    & Acc. & Format & Fact & Cons. & & & & \\
    \midrule
    & & & & & 61.21 & 37.32 & 66.91 & 47.91 \\
    \midrule
    $\checkmark$ & & & & & 87.45 & 69.75 & 93.44 & 79.88 \\
    \midrule
    & $\checkmark$ & $\checkmark$ & $\checkmark$ & $\checkmark$ & 82.95 & 67.94 & 67.62 & 67.77 \\
    \midrule
    $\checkmark$ & $\checkmark$ & $\checkmark$ & & & 94.84 & 91.82 & 89.11 & 90.44 \\
    $\checkmark$ & $\checkmark$ & $\checkmark$ & $\checkmark$ & & 96.77 & 94.74 & 93.25 & 93.99 \\
    \midrule
    $\checkmark$ & $\checkmark$ & $\checkmark$ & $\checkmark$ & $\checkmark$ & \cellcolor{gray!20}97.63 & \cellcolor{gray!20}95.62 & \cellcolor{gray!20}95.48 & \cellcolor{gray!20}95.55 \\ 
    \bottomrule
    \end{tabular}
}
\end{sc}
\caption{\textbf{Ablation studies of the training strategies and rewards on the IU-Xray dataset.}
PL and Cons. stand for Pseudo-Label and Consistency, respectively.}

\label{tab:comb}
\end{table}
\begin{table}[!th]
\centering
\begin{sc}
\resizebox{.95\linewidth}{!}{
\begin{tabular}{lcccc}
    \toprule
    \textbf{Models} & \textbf{Acc.} & \textbf{Pre.} & \textbf{Rec.} & \textbf{F1} \\
    \midrule
    Base VLM~\cite{qwen25-vl25_alibaba} & 61.21 & 37.32 & 66.91 & 47.91 \\
    \midrule
    {RULE~\cite{rule24emnlp_unc} \textbf{(Ours)}} & \cellcolor{gray!20}97.63 & \cellcolor{gray!20}95.62 & \cellcolor{gray!20}95.48 & \cellcolor{gray!20}95.55 \\ 
    {\color{gray!20}Human} & 98.64 & 97.96 & 96.97 & 97.46 \\
    \bottomrule
    \end{tabular}
}
\end{sc}
\caption{\textbf{The choice study of pseudo-labels for SFT.}}
\vspace{-0.3cm}
\label{tab:pls}
\end{table}

Training on medical data is crucial for achieving strong performance in this specialized domain. As shown in \cref{tab:comb}, the base generalist model \textsc{Qwen2.5-VL-3B} lacks sufficient medical expertise and frequently produces false-positive misdiagnoses, yielding a low precision of $37.32$.
{Supervised learning with pseudo-labels} (\cref{sec:sft}) and reinforcement learning with reward signals (\cref{sec:rl}) significantly improve the \textsc{Qwen2.5-VL-3B} baseline.

Interestingly, these two approaches lead to distinct model behaviors.
Supervised fine-tuning (\textsc{SFT}) enables the model to effectively identify positive diseases by imitating reliable and factual pseudo-diagnoses~\cite{rule24emnlp_unc}, achieving a recall gain of $39.7\%$ and reaching up to $0.9344$ (see the 2\textsuperscript{nd} row).
{In contrast, reinforcement learning, where we adopt the most effective variant {GRPO}~\cite{grpo24_ds} promotes conservative predictions, as evidenced by the similar precision, recall, and F1 scores of approximately $0.67$ (see the 3\textsuperscript{rd} row).}

The combination enhances fact-grounded reasoning~\cite{ds-r125_ds}, yielding an additional $11.6\%$ gain and pushing accuracy to $0.9763$.
Notably, despite imperfections in the factual pseudo-labels, their integration within the GRPO framework rivals results achieved via human-annotated SFT, underscoring the robustness and scalability of our approach (\cref{tab:pls}).
\vspace{-0.3cm}
\subsubsection{Rewards}

Reward design is widely recognized as a critical factor influencing the performance of reinforcement learning.
As shown in the 4\textsuperscript{th} row of \cref{tab:comb}, building on pseudo-label SFT, rewarding only the \emph{accuracy} of the binary diagnosis and a simple output \emph{format} enables the {VLM} to capture typical disease characteristics, yielding a strong $+10.56$ F1 score gain.
This highlights the effectiveness of RL post-training for medical applications.
The \emph{fact} reward further enriches the outputs with domain-specific terminology, whereas the \emph{consistency} reward enhances logical coherence by aligning intermediate reasoning with final answers.
Each contributes further improvements of $+3.55$ and $+1.56$, respectively, underscoring their complementary roles in strengthening factual reasoning and the structured generation of responses.

\vspace{-0.3cm}
\section{Discussions \& Conclusions}

\logopic\ours\ establishes a robust two-stage framework for factual medical reasoning, integrating pseudo-label generation and GRPO-based reinforcement learning.  
It delivers substantial gains in factuality and reliability across diverse medical QA benchmarks, highlighting the value of combining external knowledge with adaptive-policy optimization.  
The experiments also reveal that SFT initialization and reward design are critical for ensuring stable and effective training.  
Despite these advances, challenges remain in scaling complex real-world clinical scenarios and ensuring robustness against rare or ambiguous cases.  
Future research should explore richer reward functions, video reasoning, and deployment challenges, including safety and fairness.  
We believe that continued innovation in knowledge integration and RL incentivization will be promising for advancing trustworthy and generalizable medical AI.

\bibliographystyle{IEEEbib}
\bibliography{strings,refs}

\end{document}